\documentclass{article} 
\pdfoutput=1
\usepackage{nips13submit_e,times}
\usepackage{hyperref}
\usepackage{url}
\usepackage{graphicx}
\usepackage{LIreduced}
\newcommand{\cf}[1]{\mbox{$\mathit{#1}$}}   

\title{Learning Type-Driven Tensor-Based Meaning Representations}

\author{
Tamara Polajnar and Luana F\v{a}g\v{a}ra\c{s}an and  Stephen Clark \\
University of Cambridge\\
Computer Laboratory\\
\texttt{first.last@cl.cam.ac.uk} 
}

\nipsfinalcopy 

\begin{document}

\maketitle

\begin{abstract}

This paper investigates the learning of 3rd-order tensors representing
the semantics of transitive verbs. The meaning representations are
part of a type-driven tensor-based semantic framework, from the newly emerging field of compositional distributional
semantics. Standard techniques from the neural networks literature are
used to learn the tensors, which are tested on a selectional
preference-style task with a simple 2-dimensional sentence
space. Promising results are obtained against a competitive
corpus-based baseline. We argue that extending this work beyond
transitive verbs, and to higher-dimensional sentence spaces, is an
interesting and challenging problem for the machine
learning community to consider.

\end{abstract}

\section{Introduction}

An emerging subfield of natural language processing and computational
linguistics is concerned with learning compositional distributional
representations of meaning
\cite{Mitchell:Lapata:08,BaroniEMNLP10,LambekFest,gref:emnlp11,clarke:12,socher12,clark:13}.
The advantage of such representations lies in their potential to
combine the benefits of distributional approachs to word meaning
\cite{schuetze98,turneyPantel} with the more traditional compositional
methods from formal semantics \cite{Montague}.
Distributional representations have the properties of robustness,
learnability from data, ease of handling ambiguity, and the ability to
represent gradations of meaning; whereas compositional models handle
the unbounded nature of natural language, as well as providing
established accounts of logical words, quantification, and inference.

One promising approach which attempts to combine elements of
compositional and distributional semantics is by Coecke et
al. \cite{LambekFest}. The underlying idea is to take the type-driven
approach from formal semantics --- in particular the idea that the
meanings of complex grammatical types should be represented as
functions --- and apply it to distributional representations. Since
the mathematics of distributional semantics is provided by linear
algebra, a natural set of functions to consider is the set of linear
maps. Coecke et al. recognize that there is a natural correspondence
from complex grammatical types to tensors (multi-linear maps), so that
the meaning of an adjective, for example, is represented by a matrix
(a 2nd-order tensor)\footnote{This same insight lies behind the work
  of Baroni and Zamparelli \cite{BaroniEMNLP10}.} and the meaning of a
transitive verb is represented by a 3rd-order tensor. Coecke et
al. use the grammar of pregroups as the syntactic machinery to
construct distributional meaning representations, since both pregroups
and vector spaces can be seen as examples of the same abstract
structure, which leads to a particularly clean mathematical
description of the compositional process. However, the approach
applies more generally, for example to other forms of categorial
grammar, such as Combinatory Categorial Grammar \cite{steedman:2000},
and also to phrase-structure grammars in a way that a formal linguist
would recognize \cite{baroni:frege}. Clark \cite{clark:13} provides a
description of the tensor-based framework aimed more at computational
linguists, relying only on the mathematics of multi-linear algebra
rather than the category theory used in
\cite{LambekFest}. Section~\ref{sec:types_to_tensors} repeats some of
this description.

A major open question associated with the tensor-based semantic
framework is how to learn the tensors representing the meanings of
words with complex types, such as verbs and adjectives. The framework
is essentially a compositional framework, providing a recipe for how
to combine distributional representations, but leaving open what the
underlying vector spaces are and how they can be acquired. One
significant challenge is an engineering one: in a wide-coverage
grammar able to handle naturally occurring text, there will be a) a
large lexicon with many word-category pairs requiring tensor
representations; and b) many higher-order tensors with large numbers
of parameters which need to be learned. In this paper we take a first
step towards learning such representations, by learning tensors for
transitive verbs.

One feature of the tensor-based framework is that it allows the
meanings of words and phrases with different basic types, for example
nouns and sentences, to live in different vector spaces; but this
means that the sentence space must be specified in advance. In this
paper we consider a simple sentence space: the ``plausibility space''
described by Clark \cite{clark:13}, represented here as a probability
distribution (and hence having only 2 dimensions). Logistic regression
is used to learn a plausibility classifier.  We begin with this simple
space since we want to see the extent to which the tensor-based
representations can be learned at all.


One goal of this paper is to introduce the problem of learning
tensor-based semantic representations to the ML
community. Current methods, for example the work of Socher
\cite{socher12}, typically use only matrix representations,
and also assume that words, phrases and sentences all live in the same
vector space. The tensor-based semantic framework is more flexible, in
that it allows different spaces for different grammatical types, which
results from it being tied more closely to a type-driven syntactic
description; however, this flexibility comes at a price, since there
are many more paramaters to learn. Various communities are beginning
to recognize the additional power that tensor representations can
provide, through the capturing of interactions that are difficult to
represent with vectors and matrices (see
e.g. \cite{Ranzato10,Sutskever,cruys:coling12}). Hierarchical
recursive structures in language potentially represent a large number
of such interactions (the obvious example for this paper being the
interaction between a transitive verb's subject and object), and
present a significant challenge for machine learning.

\section{Syntactic Types to Tensors}
\label{sec:types_to_tensors}

The syntactic type of a transitive verb in English is \cf{(S\bs
  NP)/NP} (using Steedman \cite{steedman:2000} notation), meaning that
a transitive verb is a function which takes an \cf{NP} argument to the
right, an \cf{NP} argument to the left, and results in a sentence
\cf{S}. Such {\em categories} with slashes are {\em complex
  categories}; \cf{S} and \cf{NP} are {\em basic} or {\em atomic}
categories. Interpreting such categories under the Coecke et
al. framework is straightforward. First, for each atomic category
there is a corresponding vector space; in this case the sentence space
$\mathbf{S}$ and the noun space $\mathbf{N}$.\footnote{In practice,
  for example using the CCG parser of Clark and Curran
  \cite{clark:cl07}, there will be additional atomic categories, such
  as \cf{PP}, but not many more.} Hence the meaning of a noun or noun
phrase, for example {\em people}, will be a vector in the noun space:
$\overrightarrow{\mbox{\em people}} \in \mathbf{N}$. In order to
obtain the meaning of a transitive verb, each slash is replaced with a
tensor product, so that the meaning of {\em eat}, for example, is a
3rd-order tensor: $\overline{\mbox{\em eat}} \in \mathbf{S\otimes N
  \otimes N}$. Just as in the syntactic case, the meaning of a
transitive verb is a function (a multi-linear map) which takes two
noun vectors as arguments and returns a sentence vector.

Meanings combine using {\em tensor contraction}, which can be thought
of as a multi-linear generalisation of matrix multiplication
\cite{gref_thesis}. Consider first the adjective-noun case, for
example {\em black cat}. The syntactic type of {\em black} is
\cf{N/N}; hence its meaning is a 2nd-order tensor (matrix):
$\overline{\mbox{\em black}} \in \mathbf{N \otimes N}$. In the syntax,
\cf{N/N} combines with \cf{N} using the  rule of forward
application (\cf{N/N} \cf{N} $\Rightarrow$ \cf{N}), which is an
instance of function application. Function application is also used in
the tensor-based semantics, which, for a matrix and vector argument,
corresponds to matrix multiplication.

\begin{figure}
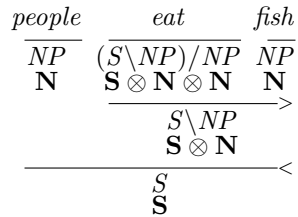


\begin{center}
\deriv{3}{
  \it people & \it eat & \it fish \\
  \uline{1}&\uline{1}&\uline{1}\\
  \it \cf{NP} &\it \cf{(S \bs NP) /NP} &\it \cf{NP}\\
  \mathbf{N} & \mathbf{S} \otimes \mathbf{N} \otimes \mathbf{N} & \mathbf{N} \\
 \ & \fapply{2} \\
  \ & \multicolumn{2}{c}{{\it \cf{S} \bs NP}}\\
  \ & \multicolumn{2}{c}{\mathbf{S} \otimes \mathbf{N}}\\
 \bapply{3} \\
 \multicolumn{3}{c}{{\it S}} \\
  \multicolumn{3}{c}{\mathbf{S}}\\
}
\end{center}

\caption{Syntactic reduction and tensor-based semantic types for a transitive verb sentence}
\vspace*{-0.3cm}
\label{fig:deriv}

\end{figure}

Figure~\ref{fig:deriv} shows how the syntactic types combine with a
transitive verb, and the corresponding tensor-based semantic
types. Note that, after the verb has combined with its object \cf{NP},
the type of the verb phrase is \cf{S\bs NP}, with a corresponding
meaning tensor (matrix) in $S \otimes N$. This matrix then combines
with the subject vector, through matrix multiplication, to give a
sentence vector.

In practice, using for example the wide-coverage grammar from CCGbank
\cite{hock:cl07}, there will be many types with more than 3 slashes,
with corresponding higher-order tensors. For example, a common
category for a preposition is the following: \cf{((S\bs NP)\bs (S\bs
  NP))/NP}, which would be assigned to {\em with} in {\em eat with a
  fork}. (The way to read the syntactic type is as follows: {\em with}
requires an \cf{NP} argument to the right -- {\em a fork} in this
example -- and then a verb phrase to the left -- {\em eat} with type
\cf{S\bs NP} -- resulting in a verb phrase.) The corresponding meaning
tensor lives in the space $S\otimes N \otimes S \otimes N \otimes N$,
i.e. a 5th-order tensor. Categories with even more slashes are not
uncommon, for example \cf{((N/N)/(N/N))/((N/N)/(N/N))}. Clearly
learning parameters for such tensors is highly challenging, and it is
likely that lower dimensional approximations will be required. We
leave investigation of such approximations to future work.

\section{Verb and Sentence Representation}

As described above, in this paper we have chosen to focus on a two-dimensional
``plausibility space'' for the meanings of sentences. One way to think
of this space is the simplest extension of truth values from the
traditional truth-theoretic account to a real-valued
setting.\footnote{We are not too concerned with the philosophical
  interpretation of these plausibility values; rather we see the
  plausibility sentence space as a useful inital testbed for the
  tensor-based semantic framework.}  We also focus on the plausibility
of transitive verb sentences with the simple {\em subject verb object}
(SVO) grammatical structure, for example {\em people eat fish} (as in
Figure~\ref{fig:deriv}). These sentences were generated automatically
by finding specific transitive verbs in a dependency-parsed corpus and
extracting the head nouns from the subject and the object (see
Section~\ref{datagen}). The nouns have atomic syntactic types and are
represented by distributional semantic vectors, built using standard
techniques \cite{turneyPantel}, while the verb is a multi-linear map
that takes in two nouns and outputs values in the plausibility space.

We define the plausibility space to have two dimensions, one
corresponding to {\em plausible} and one corresponding to {\em
  implausible}. Hence the verb tensor outputs two real values for each
subject-verb-object triple. If the vectors in noun space have
dimensionality $K$ and the sentence space has dimensionality $S$ (two
in this case), then the verb is a $K\times K\times S$ tensor. We add
some additional processing to the tensor network, following standard
practice in neural networks and following
\cite{Krishnamurthy13}, by passing the output values though a
non-linear sigmoid function, and then creating a probability
distribution over over two classes, {\em plausible ($\top$)} and {\em
  implausible ($\bot$)}, using a softmax function.



In Section~\ref{learning}, we propose a two-class logistic regression
classifier for simultaneous learning of the verb function and the
plausibility space. This method was introduced in
\cite{Krishnamurthy13}, but only implemented as a proof-of-concept
with vectors of length 2 and small, manually created datasets based on
propositional logic examples. In order to make the learning practical,
given the large numbers of contextual features in the noun vectors, we
employ a technique (described in Section~\ref{nounvecs}) that improves
low-dimensional singular value decomposition (SVD)
\cite{Deerwester90}, and thus enables us to effectively limit the
number of parameters while learning from corpus-sized data. As a
baseline we adapted a method from \cite{gref:emnlp11},
where the verb is represented as the average of the Kronecker products
of the subject and object vectors from the positive training
data. This method does not produce a plausibility space, but
plausibility of the subject-verb-object triple can be calculated using
cosine similarity (see Section~\ref{baseline}).

\subsection{Tensor learning}
\label{learning}

Following \cite{Krishnamurthy13}, we learn the tensor values as
parameters ($\mathrm{V}$) of a regression algorithm. To represent this
space as a distribution over two classes ($\top$,$\bot$) we apply a
sigmoid ($\sigma$) to restrict the output to the [0,1] range and the
softmax activation function ($g$) to balance the class probabilities.
The full parameter set which we need to optimise for is
$\mathrm{B}=\{\mathrm{V},\Theta \}$, where $\Theta=\{\theta_{\top},
\theta_{\bot}\}$ are the softmax parameters for the two classes.  For
each verb we optimise the KL-divergence $\mathcal{L}$ between the
training labels $t^i$ and classifier predictions using:
\begin{equation}
O(\mathrm{B}) = \sum_{i=1}^N \mathcal{L}\left( t^i,
g\left(\sigma\left((n_s^i)\mathrm{V}(n_o^i)^T\right),\Theta\right)\right) + \frac{\lambda}{2}||\mathrm{B}||^2
\end{equation}
where $n_s^i$ and $n_o^i$ are the subject and object of the training
instance $i\in N$. The gold-standard distribution over training labels
is defined as $(1,0)$ or $(0,1)$, depending on whether the training
instance is a positive (plausible) or negative (implausible) example.
The derivatives are obtained via the chain rule with respect to each
set of parameters and gradient descent is performed using the Adagrad
algorithm \cite{Duchi11}. Tensor contraction is implemented using the
Matlab Tensor Toolbox \cite{TTB}.

\subsection{Baseline}
\label{baseline}

The baseline is a simple corpus-driven approach of generating a matrix
from an average of Kronecker products of the subject and object
vectors from the positively labelled subset of the training data
\cite{gref:emnlp11}, for each verb. The intuition is that the matrix
for each verb represents an average of the pairwise contextual
features of a typical subject and object (as extracted from instances
of the verb). For example, the matrix for {\em eat} may have a high
value for the contextual feature pair ({\em is\_human}, {\em
  is\_food}) (assuming that the features are interpretable in this
way). To determine the plausibility of a new subject-object pair for a
particular verb, we calculate the Kronecker product of the subject and
object noun vectors for this pair, and compare the resulting matrix
with the average verb matrix using cosine similarity. Intuitively, the
average verb matrix can be thought of as what the verb expects to see
in terms of the contextual features of its subject and objects, and
the cosine is determining the extent to which the particular argument
pair satisfies those expectations.  As well as being an intuitive
corpus-based method for representing the meaning of a transitive verb,
this method has also performed well experimentally
\cite{gref:emnlp11}, and hence we consider it to be a competitive
baseline.

For label prediction, the cutoff is estimated at the break-even point
of the receiver operator characteristic (ROC) generated by testing the
positive and negative examples of the training data against the
learned average matrix.\footnote{The break-even point is when the true
  positive rate is equal to the false positive rate.}  In practice it
would be more accurate to estimate the cutoff on a validation dataset,
but some of the verbs have so few training instances that this was not
possible.  


\section{Data}
\label{datagen}

To train a classifier for each verb, a dataset of positive and
negative examples is required.  While we can consider
subject-verb-object triples that naturally occur in corpus data as
positive, a technique for generating pseudo-negative examples is
needed, which is described below.


\subsection{Training examples}

In order to generate training data we made use of two large corpora:
the Google Syntactic N-grams (GSN) \cite{Goldberg13} and the Wikipedia
October 2013 dump. The Wikipedia corpus consists of the textual
content tokenised using the Stanford NLP
tools\footnote{http://nlp.stanford.edu/software/index.shtml} and
parsed and lemmatised using the C\&C parser and the Morpha lemmatiser
\cite{clark:cl07,carroll:01}.

We first chose transitive verbs with different concreteness scores
\cite{brysbaert2013concreteness} and frequencies, in order to obtain a
variety of verb types. Then the positive SVO examples were extracted
from the GSN corpus.  More
precisely, we extracted all distinct syntactic trigrams of the form
\emph{nsubj ROOT dobj}, where the root of the phrase was one of our
target verbs. We lemmatised the words using the
NLTK\footnote{http://nltk.org/} lemmatiser and filtered these examples
to retain only the ones that contain nouns that also occur in
Wikipedia, obtaining the counts reported in Table \ref{table:verbs}.

\begin{table}[t]
{\sc \footnotesize
\begin{center}
\begin{tabular}{ l | l l l }
\hline
 \multicolumn{1}{c}{\bf Verb} & \multicolumn{1}{|c}{\bf Concreteness} & \multicolumn{1}{c}{\bf \# of Positive Examples} & \multicolumn{1}{c}{\bf Frequency} \\
 \hline\hline
apply & 2.5 & 5618 & 47361762 \\ 
censor & 3 & 26 & 278525 \\ 
comb & 5 & 164 & 644447 \\
depose & 2.5 & 118 & 874463 \\
eat & 4.44 & 5067 & 26396728 \\
idealize & 1.17 & 99 & 485580 \\
incubate & 3.5 & 82 & 833621 \\
justify & 1.45 & 5636 & 10517616 \\ 
reduce & 2 & 26917 & 40336784 \\ 
wipe & 4 & 1090 & 6348595 \\ 
\hline
\end{tabular}
\end{center}
}
\caption{The 10 chosen verbs together with their concreteness scores. The number of positive SVO examples was capped at 2000. {\bf Frequency} is the frequency of the verb in the GSN corpus. }
\label{table:verbs}
\end{table}

For every positive (plausible) training example, we constructed a
negative (implausible) one by replacing both the subject and the
object with a \emph{confounder}, using a standard technique from the
selection preferences literature \cite{chambers2010improving}. A
confounder was generated by choosing a random noun from the same
frequency bucket as the original noun. Frequency buckets of size 10
were constructed by collecting noun frequency counts from the
Wikipedia corpus. Table \ref{table:examples} presents a few pairs of
positive and negative training examples.

\begin{table}[t]
{\sc \footnotesize
\begin{center}
\begin{tabular}{ l l }
\hline
 \multicolumn{1}{c}{\bf Positive} & \multicolumn{1}{c}{\bf Negative}  \\
\hline\hline
{\em court} apply {\em law} & {\em plan} apply {\em title} \\
{\em woman} comb {\em hair} & {\em role} comb {\em guitarist} \\
{\em animal} eat {\em plant} & {\em mountain} eat {\em product} \\
\hline
\end{tabular}
\end{center}
}\
\vspace*{-0.3cm}
\caption{Some example training instances}
\label{table:examples}
\vspace*{-0.3cm}
\end{table}

\subsection{Noun representation}
\label{nounvecs}

Distributional semantic models \cite{turneyPantel} encode word meaning
in a vector format by counting co-occurrences with other words within
a specified context, which can be defined in many ways, for example as
a whole document, an N-word window, or a grammatical relation. In this
paper we use sentence boundaries to define context windows.  To
generate noun context vectors, the Wikipedia corpus described above is
scanned for the nouns that appear in the training data and the number
of times a context word ($c_j$) occurs within the same sentence as the
target noun ($w_i$) is recorded in the vector representing that
noun. The context words are the top 10,000 most frequent lemmatised
words in the whole corpus excluding stopwords. The raw co-occurrence
counts are re-weighted using the standard tTest weighting scheme,
where $f_{w_ic_j}$ is the number of times target noun $w_i$ occurs
with context word $c_j$:

\begin{equation}
tTest(\vec{w_i}, c_j) = \frac{p(w_i, c_j)-p(w_i)p(c_j)}{\sqrt{p(w_i)p(c_j)}}
\end{equation}

where  {\small $p(w_i)=\frac{\sum_jf_{w_ic_j}}{\sum_k\sum_lf_{w_kc_l}}$},  {\small $p(c_j)=\frac{\sum_if_{w_ic_j}}{\sum_k\sum_lf_{w_kc_l}}$}, and {\small $p(w_i, c_j)=\frac{f_{w_ic_j}}{\sum_k\sum_lf_{w_kc_l}}$}.

Using all 10,000 context words would result in a large number of
parameters for each verb tensor, and so we apply the following
dimensionality reduction technique which makes training tractable.
Considering tTest values as a ranking function, we choose the top $N$
highest ranked context words for each noun. The value $N$ is chosen by
testing on the development subset of the MEN dataset (MENdev), a
standard dataset for evaluating the quality of semantic vectors
\cite{Bruni12}.\footnote{The MEN dataset contains 3000 word pairs that
  were judged for similarity by human annotators. Of that 2000 are in
  the development subset and the remaining 1000 are used as test
  data.}  
The tTest weights span the range $[-1, 1]$, but are
generally tightly concentrated around zero. Hence an additional
technique we use is to spread the range using row normalisation:
$\vec{w} := \frac{\vec{w}}{||\vec{w}||_2}$. Hence each noun vector now
contains only $N$ non-zero values, where each value is a (weighted,
normalised) co-occurrence frequency.

\begin{figure}[t]
\begin{center}
\includegraphics[width=0.50\textwidth]{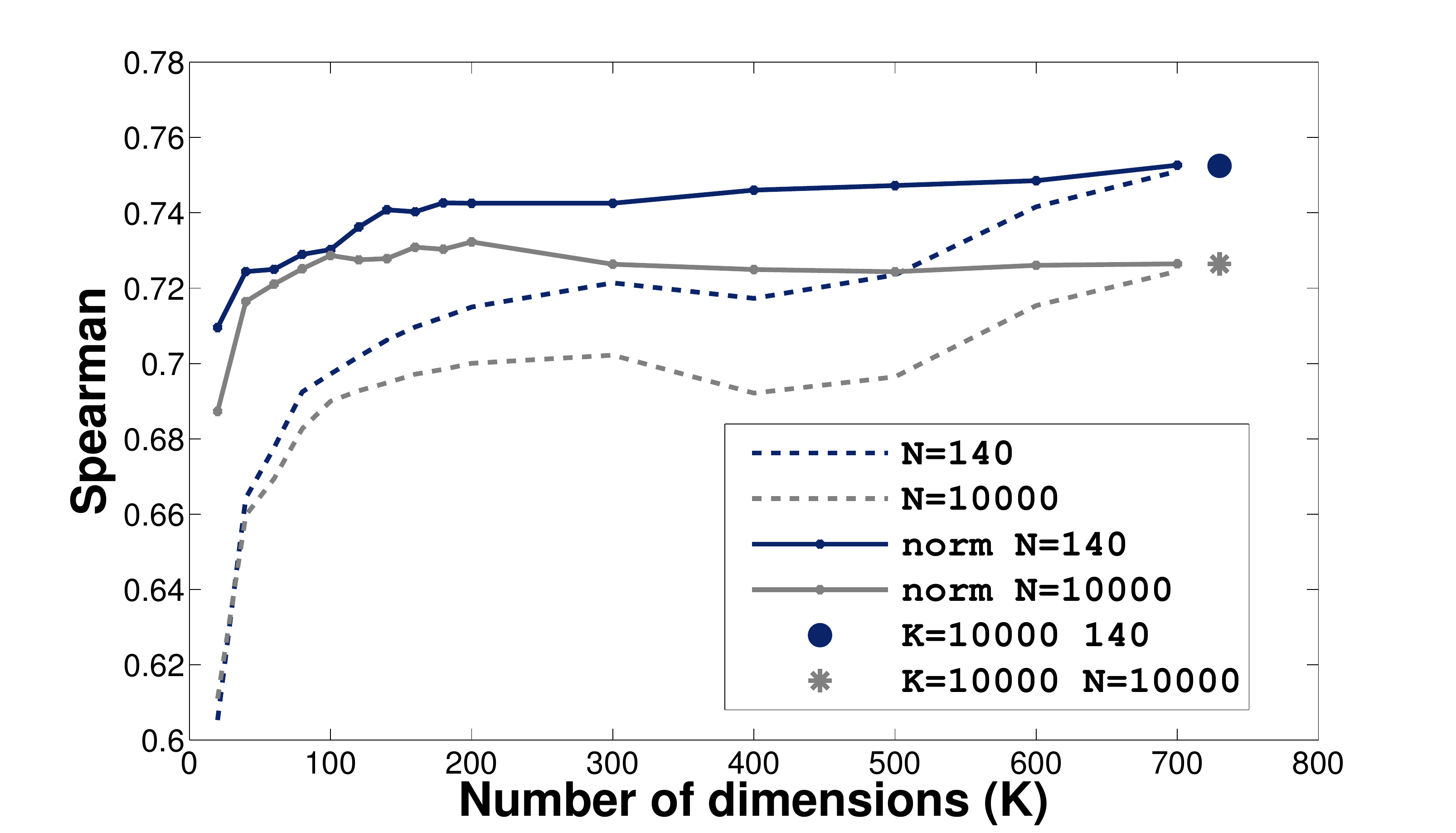}
\end{center}
\caption{Vectors tuned for sparseness (dark) consistently produce equal or better dimensionality reductions (MENdev). The solid lines show improvement in lower dimensional representations of SVD when dimensionality reduction is applied after normalisation.  }
\vspace*{-0.3cm}
\label{dimred}

\end{figure}

Finally, placing each noun vector as a row in a matrix results in a
noun-context co-occurrence matrix.  Singular value decomposition (SVD)
is applied to this matrix, with 20 latent dimensions. Applying SVD to
such a matrix is a standard technique for removing noise and
uncovering the latent semantic dimensions in the data, and has been
reported to improve performance on a number of semantic similarity
tasks \cite{turneyPantel}.  Together these two simple techniques
\cite{Polajnar14} markedly improve the performance of SVD on smaller
dimensions ($K$) on the MENdev set (see Figure~\ref{dimred}), and
enable us to train the verb tensors using 20-dimensional noun
vectors. On an older, highly reported dataset of 353 word pairs
\cite{finkelstein:02} our vectors achieve the Spearman correlation of
0.63 without context selection and normalisation, and 0.60 with only
20 dimensions after these techniques have been applied.On MENtest we
get 0.73 and 0.71, respectively.

\section{Experiments}

We conducted three experiments. The first used all of the available
training examples in 5 repetitions of a 2-fold cross-validation
(5x2cv) experiment to evaluate the peak performance for each of the
verbs (Table~\ref{table:fullcv}). The verbs with many subject-object
pairs were capped at 4,000 instances (the 2,000 most frequent positive
pairs and 2,000 pseudo-negative). We compared the performance of the
{\em baseline} and {\em tensor learning} methods on 20 and 40
dimensional vectors.\footnote{We also implemented and experimented with a {\em matrix} method, which outputs a sigmoid transformed single plausibility value instead of the overparameterised 2-value softmax vector. This method performed worse than baseline and was thus left out of this paper.}

The performance was evaluated using the area
under the ROC (AUC) and the F$_1$ measure (based on precision and
recall over the plausible class). The AUC evaluates whether a method
is ranking positive examples above negative ones, regardless of the
class cutoff value.  F$_1$ shows how accurately a method assigns the
correct class label. Since the baseline uses {\em ad hoc} class
assignment, AUC is the more fair measure.



\begin{figure}
\begin{center}
\includegraphics[width=0.45\textwidth]{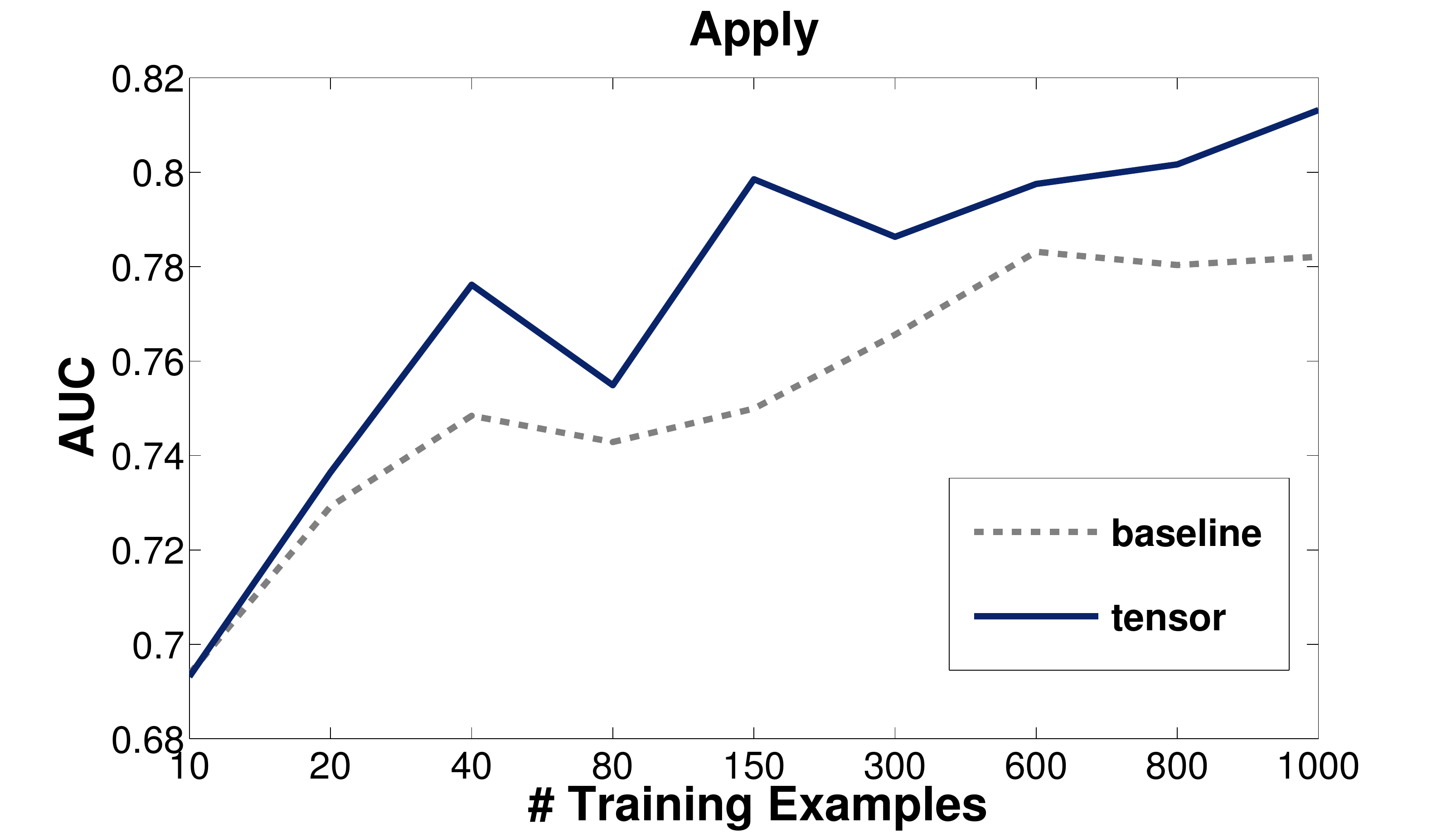}
\includegraphics[width=0.45\textwidth]{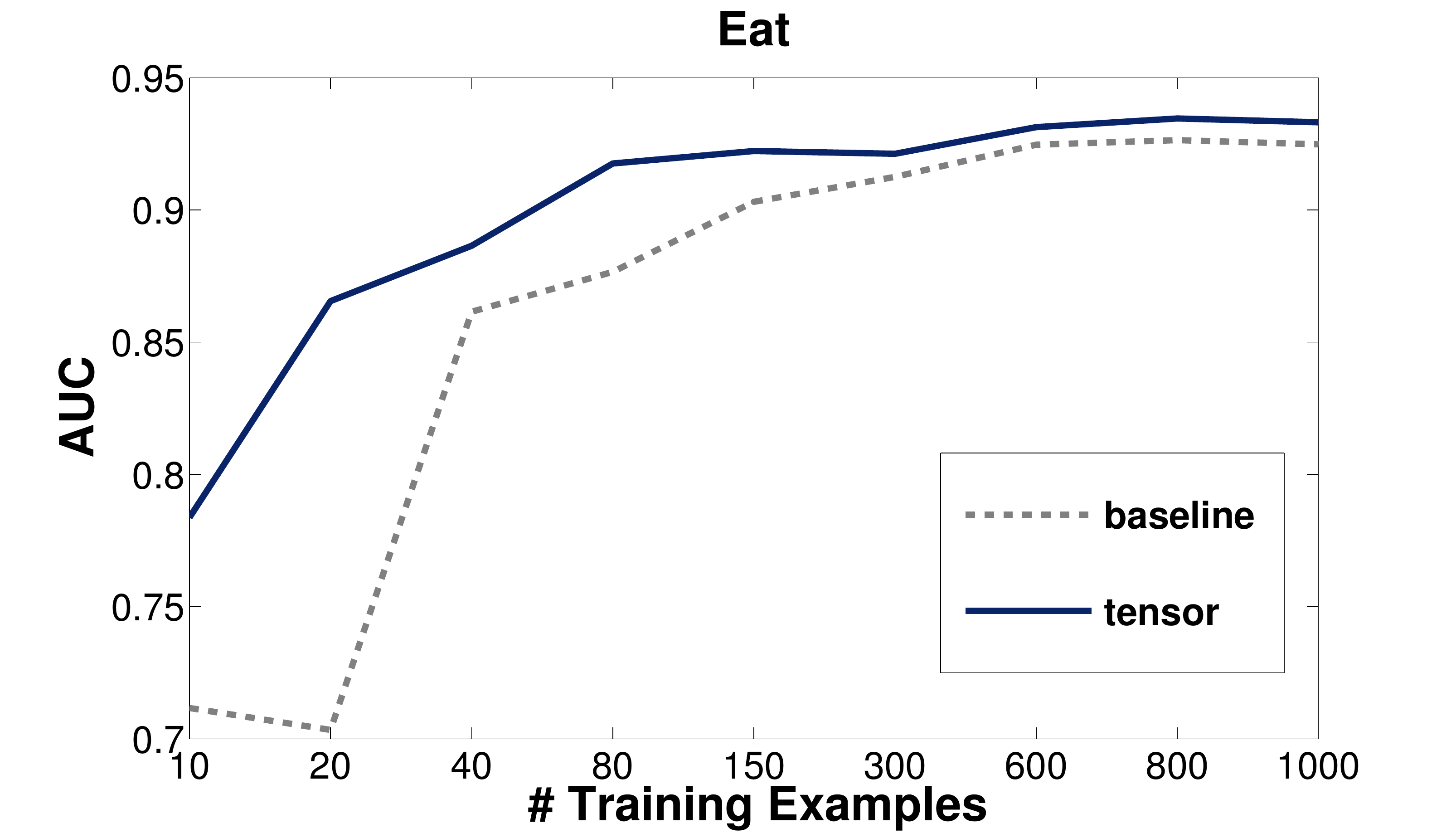}
\includegraphics[width=0.45\textwidth]{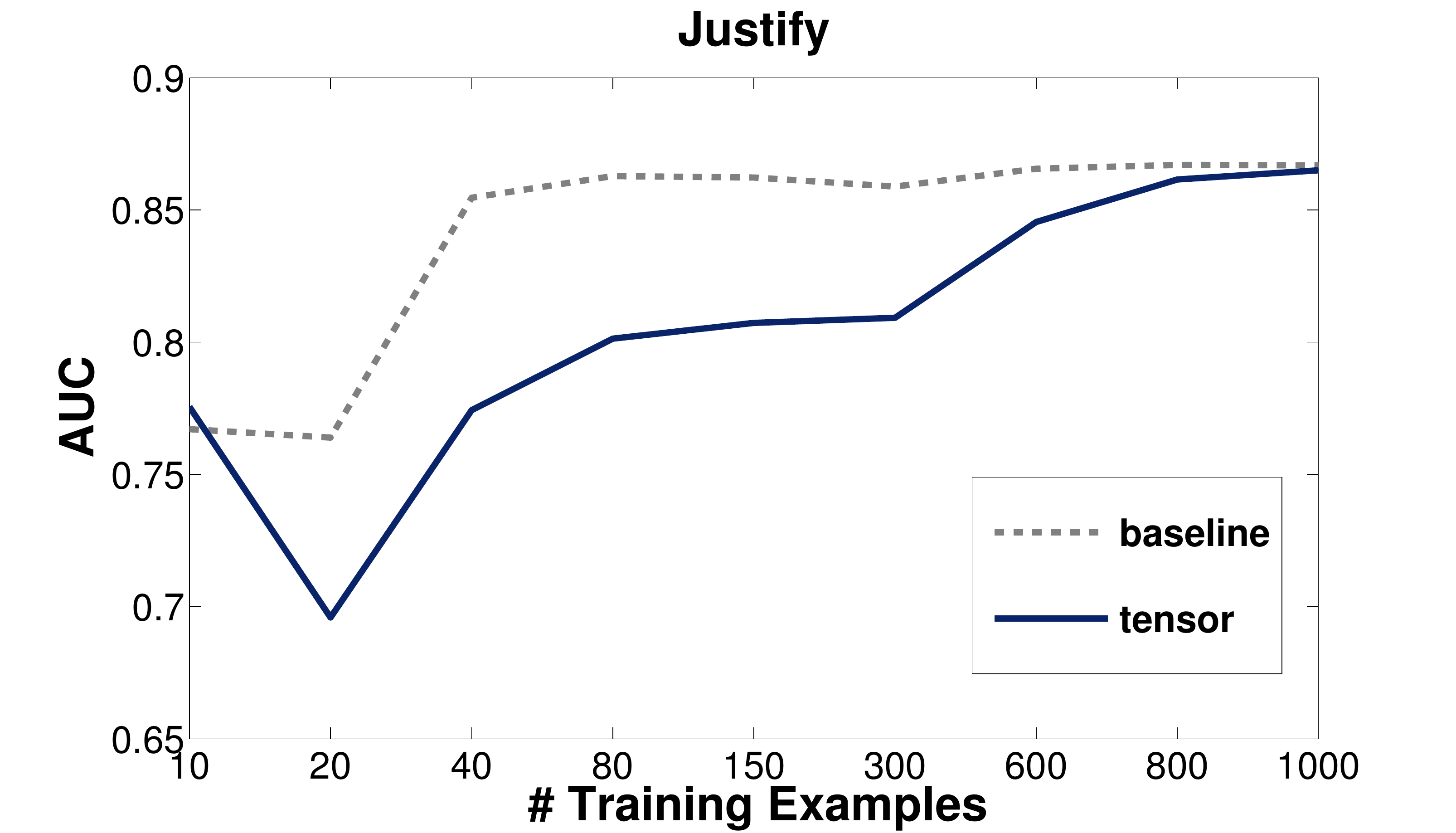}
\includegraphics[width=0.45\textwidth]{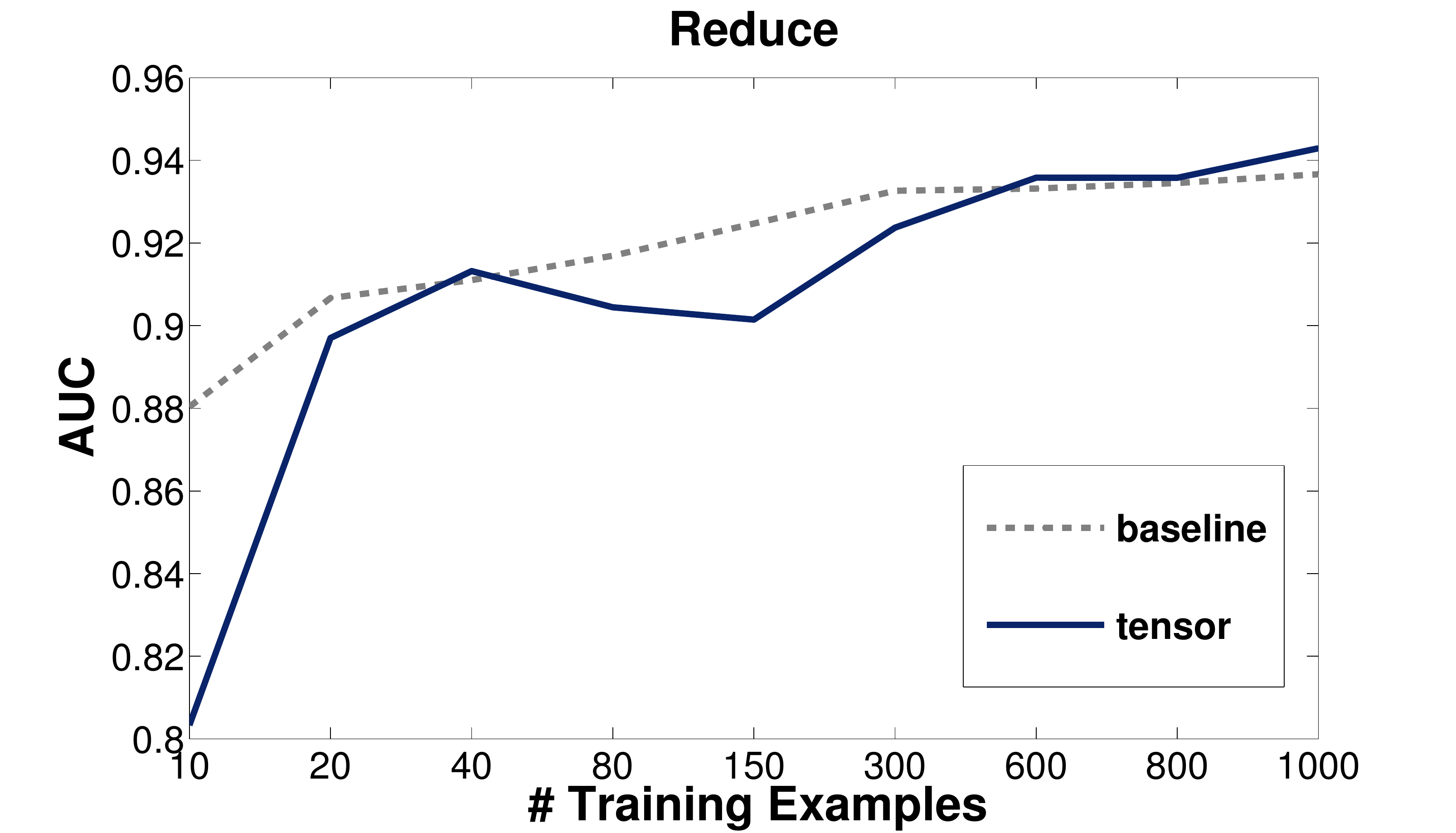}
\end{center}
\caption{Comparison of baseline (dashed) and tensor learning (full) methods as the number of training instances increases. }
\label{trsize}
\vspace*{-0.3cm}
\end{figure}

In the second experiment, we repeated the 5x2cv with datasets of 52
training points for each verb, as this is the size of the smallest
dataset of the verb {\sc censor} (Table~\ref{table:smallcv}). The
points were randomly sampled from the datasets used in the first
experiment. Finally, the four verbs with the largest datasets were
used to examine how the performance of the methods change as the
amount of training data increases. The 4,000 training samples were
randomised and half was used for testing. We sampled between 10 and
1000 training triples from the other half (Figure~\ref{trsize}).


\begin{table}[t]
{\sc \footnotesize
\begin{center}
\begin{tabular}{lll|ll}
\multicolumn{1}{c}{\bf Verb} &\multicolumn{2}{c}{\bf Vectors $K=20$}  &\multicolumn{2}{c}{\bf Vectors $K=40$}
\\ \hline \\
 &\multicolumn{1}{c}{\bf Baseline} &\multicolumn{1}{c}{\bf Tensor}  &\multicolumn{1}{c}{\bf Baseline} &\multicolumn{1}{c}{\bf Tensor}    
\\ \hline
\multicolumn{5}{c}{\bf AUC} \\ \hline
apply & 78.67 $\pm$ 0.81 & {\bf 84.81 $\pm$ 0.69}$\dagger$ & 81.46 $\pm$ 0.58 & {\bf 85.68 $\pm$ 0.72}$\dagger$\\
censor & {\bf 89.79 $\pm$ 7.52} & 82.01 $\pm$ 6.02& {\bf 85.54 $\pm$ 9.04} & 79.40 $\pm$ 8.23 \\
comb & 82.93 $\pm$ 1.70 & {\bf 87.68 $\pm$ 2.19} & 85.65 $\pm$ 2.13 & {\bf 89.41 $\pm$ 2.33} \\
depose & {\bf 92.78 $\pm$ 1.44} & 91.10 $\pm$ 1.85 & {\bf 94.44 $\pm$ 1.16} & 92.70 $\pm$ 1.48 \\
eat & 92.99 $\pm$ 0.43 & {\bf 94.01 $\pm$ 0.68} &  93.81 $\pm$ 0.45 & {\bf 94.62 $\pm$ 0.37} \\
idealize & {\bf 75.18 $\pm$ 7.48} & 69.52 $\pm$ 2.83 & {\bf 75.84 $\pm$ 6.85} & 69.56 $\pm$ 4.52 \\
incubate & 79.70 $\pm$ 4.85 & {\bf 84.94 $\pm$ 2.64}& 85.53 $\pm$ 2.78 & {\bf 89.33 $\pm$ 2.52} \\
justify & 87.32 $\pm$ 0.85 & {\bf 88.21 $\pm$ 0.61} & {\bf 88.70 $\pm$ 0.62} & 85.27 $\pm$ 0.93 \\
reduce & 94.24 $\pm$ 0.38 & {\bf 95.03 $\pm$ 0.46} & 95.48 $\pm$ 0.48 & {\bf 96.13 $\pm$ 0.37} \\
wipe & 80.53 $\pm$ 1.04 & {\bf 82.00 $\pm$ 1.18}  & 84.47 $\pm$ 0.98 & {\bf 85.19 $\pm$ 1.16} \\

\multicolumn{5}{c}{\bf F$_1$} \\ \hline 
apply & 64.24 $\pm$ 13.90 & {\bf 77.37 $\pm$ 0.57} & 64.00 $\pm$ 16.48 & {\bf 79.27 $\pm$ 1.03} \\
censor & 56.12 $\pm$ 34.37 & {\bf 74.62 $\pm$ 7.22} & 47.93 $\pm$ 31.08 & {\bf 70.66 $\pm$ 10.83} \\
comb & 52.38 $\pm$ 31.35 & {\bf 80.19 $\pm$ 3.36} & 45.02 $\pm$ 34.25 & {\bf 81.15 $\pm$ 2.59} \\
depose & 56.84 $\pm$ 31.87 & {\bf 84.50 $\pm$ 1.73} & 54.77 $\pm$ 38.04 & {\bf 84.60 $\pm$ 2.66} \\
eat & 54.03 $\pm$ 33.38 & {\bf 87.78 $\pm$ 0.74} & 52.45 $\pm$ 27.68 & {\bf 88.91 $\pm$ 0.54} \\
idealize & 49.03 $\pm$ 24.74 & {\bf 53.61 $\pm$ 28.43} & 48.28 $\pm$ 23.41 & {\bf 66.53 $\pm$ 4.69} \\
incubate & 55.42 $\pm$ 29.80 & {\bf 77.81 $\pm$ 3.95}& 50.84 $\pm$ 37.99 & {\bf 80.30 $\pm$ 5.51} \\
justify & 69.70 $\pm$ 14.41 & {\bf 81.44 $\pm$ 0.69} & 73.71 $\pm$ 8.74 & {\bf 79.73 $\pm$ 0.94} \\
reduce & 77.26 $\pm$ 6.99 & {\bf 89.06 $\pm$ 0.55}  & 71.24 $\pm$ 17.23 & {\bf 91.24 $\pm$ 0.57} \\
wipe & 55.94 $\pm$ 27.64 & {\bf 75.77 $\pm$ 1.21} & 47.62 $\pm$ 33.63 & {\bf 78.57 $\pm$ 0.73} \\
\end{tabular}
\end{center}
}
\caption{Full-size data cross-validation results with standard deviation. Bold indicates that the method
  performs better, and $\dagger$ that the result is significant according to the 5x2cv F-test \cite{Ulas12}.}
\label{table:fullcv}
\end{table}

\begin{table}[t]
{\sc \footnotesize
\begin{center}
\begin{tabular}{lll}
\multicolumn{1}{c}{\bf Verb} &\multicolumn{1}{c}{\bf Baseline} &\multicolumn{1}{c}{\bf Tensor}
\\ \hline \\
apply & 67.03 $\pm$ 8.86 &  {\bf 83.24 $\pm$ 4.06} \\
censor & 83.55 $\pm$ 6.81 & 83.93 $\pm$ 3.96 \\
comb & 71.87 $\pm$ 11.48 & {\bf 81.26 $\pm$ 7.28} \\
depose & 92.74 $\pm$ 3.45 & {\bf 95.84 $\pm$ 2.86} \\
eat & 73.96 $\pm$ 3.47 & {\bf 90.14 $\pm$ 3.02}$\dagger$ \\
idealize & {\bf 66.72 $\pm$ 4.45} & 52.35 $\pm$ 7.54 \\
incubate & 51.52 $\pm$ 8.14 & {\bf 79.75 $\pm$ 7.01}$\dagger$ \\
justify & 72.36 $\pm$ 12.66 & {\bf 75.44 $\pm$ 9.25}\\
reduce & 79.69 $\pm$ 5.24 & {\bf 91.25 $\pm$ 4.33} \\
wipe & 76.70 $\pm$ 8.14 & 75.24 $\pm$ 5.70 \\
\end{tabular}
\end{center}
}
\caption{Small data (26 positive + 26 negative per verb)
  cross-validation results show AUC with standard deviation. The $\dagger$ indicates statistically significant results.}
\label{table:smallcv}
\vspace*{-0.3cm}
\end{table}



\subsection{Analysis}

In general the tensor learning algorithm learns more effectively and
with smaller variance than the baseline, particularly from the smaller
dimensional noun vectors. The F$_1$ scores indicate that learning is
necessary for accurate classification while the baseline AUC results
show that in principle only positive examples are necessary (since the
baseline only sees positive examples). Analysis of errors shows that
the baseline method mostly generates false negative errors
(i.e. predicting implausible when the gold standard label is
plausible), particularly on triples that contain nouns that have not
been seen with the verb in the training data, which indicates that the
baseline may not adequately generalise over the latent dimensions from
the SVD.  In contrast, tensor learning (TL) produces almost equal
numbers of false positives and false negatives, but sometimes produces
false negatives with some low frequency nouns (e.g. {\em bourgeoisie
  idealize work}), presumably because there is not enough information
in the noun vector to decide on the correct class. TL also produces
some false positive errors when either of the nouns is plausible (but
the triple is implausible), which would suggest results may be
improved by training with data where only one noun is confounded or
treating negative data as possibly positive \cite{Lee03}.

Both the full data and small data experiments indicate that {\sc
  idealize} is the most difficult verb to learn. It has the twin
properties of low frequency and low concreteness. In addition, it is
likely to have low selectional preference strength, not selecting
strongly for the semantic types of its arguments.  Both verb frequency
and concreteness have positive Spearman correlation with the TL AUC
values from Tables~\ref{table:fullcv} and
~\ref{table:smallcv}. Frequency has much stronger correlation
(Table~\ref{table:fullcv}:0.53, Table~\ref{table:smallcv}:0.31) than
concreteness (Table~\ref{table:fullcv}:0.14,
Table~\ref{table:smallcv}:0.08), even when all datasets are reduced to
the same number of examples. This is probably due to the fact that the
more frequent verbs occur in more frequent triples, which are likely
to contain highly frequent nouns, and hence have higher quality noun
vectors. However, if we just consider the most frequent verbs
(Figure~\ref{trsize}) we can see that {\sc eat}, which has the highest
concreteness (4.44), provides a much smoother learning curve and
asymptotes quicker than the less concrete verbs {\sc apply} (2.4),
{\sc reduce} (2), and {\sc justify} (1.45). From this brief analysis,
we hypothesise that noun frequency, verb concreteness, and selectional
preference strength of the verb for its arguments all influence the
quality of the learned representation.

\section{Conclusion}

In this paper we have investigated learning 3rd-order tensors to
represent the semantics of transitive verbs, with a 2-dimensional
``plausibility'' sentence space. There are obvious connections with
the large literature on selectional preference learning (see
e.g. \cite{diarmuid10} for a recent paper); however, our goal is not
to contribute to that literature, but rather to use a selectional
preference task as a first corpus-driven test of the type-driven
tensor-based semantic framework of Coecke et al., as well as introduce
this framework to the machine learning community.

We have shown that standard techniques from the neural networks
literature can be effectively applied to learning 3rd-order tensors
from corpus data, with our results showing positive trends compared to
a competitive corpus-based baseline. There is much work to be done in
extending the techniques in this paper, both in terms of a
higher-dimensional, potentially more structured, sentence space, and
in terms of incorporating the many syntactic types making up a
wide-coverage grammar. Since many of these types require higher order
tensors than 3rd-order, we suggest that tensor decomposition
techniques are likely to be necessary for practical performance.


\subsubsection*{Acknowledgments}
\small

Tamara Polajnar is supported by ERC Starting Grant DisCoTex
(306920). Stephen Clark is supported by ERC Starting Grant DisCoTex
 and EPSRC grant EP/I037512/1. Luana F\v{a}g\v{a}ra\c{s}an is
supported by an EPSRC Doctoral Training Partnership award. Thanks to
Laura Rimell and Jean Maillard for helpful discussion.


{\small \bibliographystyle{plain}
\bibliography{refs,bib}
}

\end{document}